\documentclass{bmvc2k}

\title{C-DLinkNet: Considering Multi-Level Semantic Features for Human Parsing}

\addauthor{Yu Lu}{aniki@bupt.edu.cn}{1}
\addauthor{Muyan Feng}{mayanfeng@bupt.edu.cn}{1}
\addauthor{Ming Wu}{wuming@bupt.edu.cn}{1}
\addauthor{Chuang Zhang}{zhangchuang@bupt.edu.cn}{1}

\addinstitution{
 Pattern Recognition and Intelligent System Lab\\
 Beijing University of Posts and Telecommunications\\
 Beijing, China
}

\runninghead{}{}

\begin{document}

\maketitle

\begin{abstract}
   Human parsing is an essential branch of semantic segmentation, which is a fine-grained semantic segmentation task to identify the constituent parts of human. The challenge of human parsing is to extract effective semantic features to resolve deformation and multi-scale variations. In this work, we proposed an end-to-end model called C-DLinkNet based on LinkNet, which contains a new module named Smooth Module to combine the multi-level features in Decoder part. C-DLinkNet is capable of producing competitive parsing performance compared with the state-of-the-art methods with smaller input sizes and no additional information, \textit{i}.\textit{e}., achiving mIoU=53.05 on the validation set of LIP dataset.
\end{abstract}

\section{Introduction}
\label{sec:intro}
\begin{figure}[h]
   \begin{center}
      \includegraphics[width=1\linewidth]{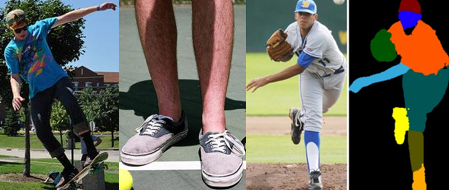}
   \end{center}
      \caption{Human parsing task}
   \label{fig:humanparsing}
   \end{figure}

Human parsing can also be considered as human semantic segmentation. This task requires a model to classify every pixel of the human body, Fig.~\ref{fig:humanparsing}. Human parsing is critical for the understanding of humans, and it advances other applications, such as dressing style recognition, human behavior recognition, and so on\cite{Zhao_2017_CVPR_Workshops}. At present, human parsing has been significantly improved as the development of deep learning and fully convolutional neural networks\cite{Long_2014,Krizhevsky2012ImageNet,Xie2016,SakLong}.
Recently, graph neural network(GNN) are used in computer vision\cite{xu2019multigraph,xu2020deep,DBLP:journals/corr/abs-1904-03751}, some people use graph convolution to capture the relationship between human parts\cite{Gong_2019_CVPR,Wang_2019_ICCV}.
In general, image semantic segmentation can be divided into two types: 1) low-level tasks, such as road extraction from satellite image, which typically lower-level features are enough to process images,2) high-level tasks, such as human parsing, which requires model to extract more semantic information. These two types of tasks require different models. For example, DLinkNet\cite{Zhou_2018_CVPR_Workshops}, which proposed for solving the road extraction problem of satellite images, uses a smaller backbone and propose Dblock to increase the receptive field. PSPNet\cite{Zhao_2017_CVPR} are proposed for a high-level task that has more bigger backbone, and the ASPP module are proposed for blend multi-level features.

Due to the potential for widespread application, research on human parsing has received increasing attention. Human parsing can be thought of as a fine-grained semantic segmentation task.  In the early years of this field, many works solved this task through Conditional Random Field(CRF) with pose estimation information\cite{Berg2012Parsing}\cite{Simo2014A}. Liang\cite{Liang2015Human}  provided a novel Contextualized Convolutional Neural Network architecture, which integrated the cross-layer, global image-level, within-super-pixel, and cross-super-pixel neighborhood context into a unified network. Gong\cite{Ke2017Look} explored a new self-supervising structure-sensitive learning method that does not require additional monitoring information, and derives a wealth of advanced knowledge from a global perspective and improves the analytical results. Liang \cite{Liang2018Look} proposed a novel joint human parsing and pose estimation network, which can have a  high-quality prediction of human parsing and pose estimation. Liu\cite{CE2P_2018_CVPR} identified some useful properties such as feature resolution, global context information, and edge information to get better results in human parsing task. 

Although the above work has verified the effectiveness of some modules in human segmentation, there is no in-depth analysis of the correlation and difference between common semantic segmentation tasks and human segmentation. In this paper, we try to explore the critical properties of human parsing and experiment with the validity of these properties. For one thing, we propose that the human segmentation task requires the network to extract features that are more robust to deformation and scale changes. For example, in the left part of Fig.~\ref{fig:humanparsing}, two people have different actions and different photographing angles, so the relative sizes between the various components of the human body are significantly different. For another, the human body segmentation task has higher semantic information requirements than the common semantic segmentation task. As shown in the right part of Fig.~\ref{fig:humanparsing}, the person's upper and lower parts have no obvious texture and color differences but are also split into two parts. Thus, the network should pay more attention to embedding semantic information into features.

Based on the above observations, we propose a unified model C-DLinknet based on DLinkNet, which is a modified version of LinkNet. And we experiment with our model and configurations, our main contributions can be summarized as follows:
   
\begin{itemize}
   \item[-]  We replace the central block by ASPP, which aggregates the contextual information effectively.
   \item[-]  We add supervised information to the output of each layer of Decoder, which makes these features of Decoder contains more semantic information.
   \item[-]  We propose a Smooth module which can concat all the output of Decoder into a hyper feature, which can give more information about scales.
\end{itemize}

\section{PROPOSED METHOD}
\begin{figure*}[h]
   \centering
      \includegraphics[width=0.97\textwidth]{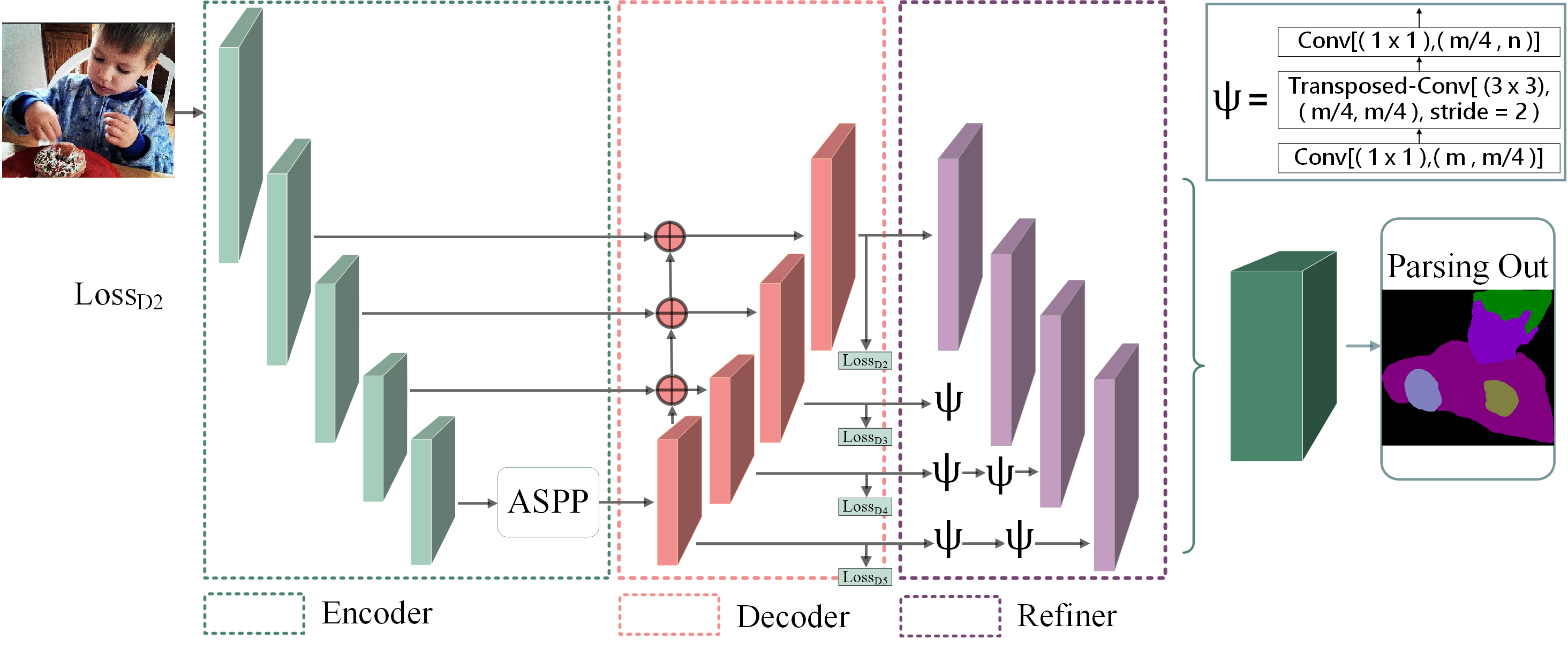}
      \caption{C-DLinkNet}
   
   \label{fig:network}
   \end{figure*}

We present a unified high-accuracy network for human parsing. This network consists of three parts: Encoder, Decoder, and Refiner. As shown in the Fig.~\ref{fig:network}, Encoder is a feature extractor that extracts robust features F. Decoder can restore the F to its original input size. Then, we improve the  Decoder by adding auxiliary loss to every Decoder layer. Also, we propose a refinement process that effectively aggregates Decoder output.

\begin{figure*}[h]
   \centering
      \includegraphics[width=\textwidth]{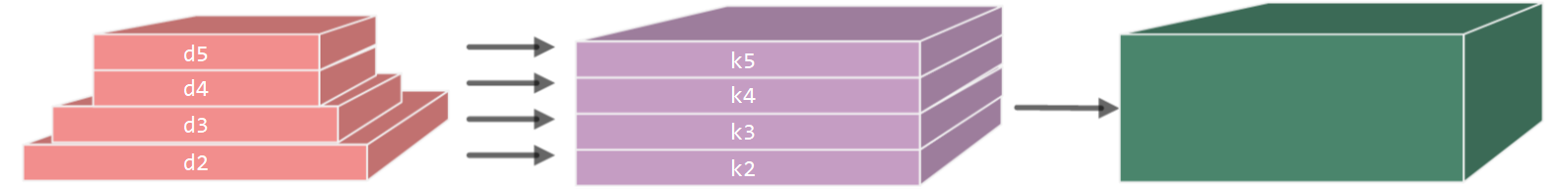}
      \caption{Smooth Module}
   
   \label{fig:decoder}
   \end{figure*}

\subsection{Encoder} 
\paragraph{Feature Extractor} We use ResNet\cite{He2015Deep} as the backbone for feature extractors. The network has 5 layers called E1, E2, E3, E4, E5. Each layer consists of different numbers of BottleNeck modules. The residual module adds up the input and output of the module in ResNet, which can effectively alleviate the vanishing gredient. Similar to DeepLab, for preserve the outsize of Encoder and retaining the receptive field, we use atrous convolution with dilate rate 2 on final E5 layer, which preserve the output size be 1/16 of the original image and finally get the feature F.
   
\paragraph{Atrous Saptial Pyramid Pooling} Global information is very helpful for fine-grained semantic segmentation. Atrous spatial pyramid pooling taking atrous convolutions with different dilation rates to get outputs with various scales. Then, these outputs effectively concat together in order to enhance outputs of Encoder.
   
Our ASPP module reduces the E5 layer output channels to 1/2 and set atrous convolutions with 12, 24, 36 dilation rates. Then, each output in the ASPP branch is reduced to 1/5 of the original input. After that, we have a bilinear interpolation to restore the features to their original size, and then we concat them together. By using a 1*1 convolution filter, we mix the features with each scale and restore features to the original number of channels before entering the module. The number of channels of feature maps F varies:
\begin{equation}
   \label{E1}
      2048\rightarrow1024\rightarrow256\rightarrow1024\rightarrow2048,
\end{equation}

There are two differences between our ASPP module and DeepLab: 
\begin{itemize}
   \item[-] To reduce the amount of computation and the number of channels, we reduce the number of the channel of input feature.
   
   \item[-] The channel numbers of output of our ASPP module are the same as the input. In this way, the output from Decoder part and Encoder part are symmetric, so it's easy for the feature maps fusion from Encoder and Decoder.
\end{itemize}

\subsection{Decoder}
The purpose of Decoder is mainly to restore the output of Encoder to the original image size, which can alleviate the problem of information loss caused by down-sampling from Encoder part. We set up 4 layers of Decoder as D5, D4, D3, and D2. We can gradually increase the size of feature maps and reduce the number of channels so that the outputs size of each Encoder layer and Decoder layer are same, so we can blend the two features by simple summation. Decoder block passes the features through a 1*1 convolution filter, so it reduces the number of channels to the original 1/4, and then it takes a deconvolution layer with stride 2 and a 1*1 convolution. This method can effectively help reduce the calculation of the model.

To add more semantic information to the output characteristics of Decoder, we add intermediate supervision to each output of Decoder. Before getting the intermediate supervision, we let the different Decoder output pass through the two-layer 3*3 convolution and get the final output by 1*1 convolution. Then we upsample this output of each Decoder layer to the ground truth size and finally get the loss, as shown in the Fig.~\ref{fig:decoder}.

Intermediate supervision can not only help features to add semantic information, but also make features more robust to different scales of objects. The loss can be formulated as:

\begin{equation}
   \label{E2}
      Loss = Loss_{refine}+0.5(Loss_{D1}+loss_{D2}+loss_{D3}+loss_{D4}),
\end{equation}


\subsection{Refiner}
Feature matters. Decoder part directly adds the output from Encoder and Decoder, which leads to the aliasing effect of the feature maps.

Inspired by HyperNet, we use the Decoder block to project the output of each layer of Decoder to the size of the feature maps from the D2 layer. After concating them together, the 3*3 convolutions blend these features and reduce the number of channels to 512. This operation effectively blends multi-scale features and increases the robustness of the network to changes in human motion. Unlike the hyper feature in HyperNet, we concat the features in Decoder. In addition to the integration of multi-scale features, it also effectively combines high-level and low-level features.

\section{Experiments}
In this section, we evaluate our method on Look Into Person (LIP) dataset with some representative methods. In addition, we also implement some comparisons with different modules by ablation study.

\subsection{Lip Dataset}
Liang proposed Look in Person (LIP benchmark dataset and related competitions)\cite{Ke2017Look}\cite{Liang2018Look}\cite{lip_web}. To further promote the frontier of semantic segmentation, and focused more on the fine-grained understanding of the human body.

The dataset is an example of a crop from the Microsoft COCO dataset. The dataset has more than 50,000 images, which are more than 30,000 training images, 10,000 validation images and 10,000 testing images. The LIP has good annotations, which are fine pixel annotations, with 19 semantic body parts and background.

LIP images are collected from real scenes and include people with challenging poses, angles, occlusions, various looks and various resolutions. As shown in Fig.~\ref{fig:humanparsing}, we can see that people's actions are various, the scenes are various and the scale changes obviously. 

\subsection{Ablation studies}
In the ablation study, we implement the proposed framework based on ResNet-101, which is pre-trained on ImageNet. The network is training on the training set and validates on the validation dataset. The input size of the network is 256*192 during training and testing. We use similar training strategies with Deeplab\cite{Chen_2018_DE}, i.e., “Poly” learning rate policy with base learning rate 0.002. We training the networks for approximately 120 epochs, and we train this model on 2 GTX1080, the batch size is 16. For data augmentation, we apply the random scaling (from 0.5 to 1.5), cropping, rotation, and left-right flipping during training and use flip for better performance. Cross-entropy is used as the loss function. 

To explore the effectiveness of the modified module, we report the performance under several variants in Tab.~\ref{tab:Ablation}. We use DLinkNet as our baseline, which proposed for the satellite image segmentation, and we can see the baseline model reaches 50.92\% accuracy. After analyzing the score of the model on each class of objects, we can observe the following problems: 1) poor recognition of small objects, such as hat, socks, and sunglasses. 2) The left and right parts of the object are more confused about the model. We replaced the D-Block module by ASPP. As shown in Tab.\ref{table:compare}, we can find it brings about 0.5\% improvements on mIoU, which demonstrates that the multi-scale context information can assist the fine-grained parsing. Particularly, it shows significant boosts on smaller objects such as glove (2.3\%), sunglasses (6.5\%). 

      \begin{table*}[htbp]
         \caption{Ablation on LIP dataset, A:ASPP Module, S:Smooth Module, L:Multi-scale Loss}
         \scalebox{0.45}{
         \begin{tabular}{|c|c|c|c|c|c|c|c|c|c|c|c|c|c|c|c|c|c|c|c|c|c|}
         \hline
         Method & bkg & hat & hair & glove & glasses & u-clo & dress & coat & sockets & pants & jsuits & scarf & skirt & face & l-arm & r-arm & l-leg & r-leg & l-shoe & r-shoe & mIoU \\
         \hline
         B&86.76&63.04&70.2&34.82&20.98&66.98&35.43&54.64&43.3&72.48&27.17&17.01&25.09&72.91&61.03&63.43&57.17&55.66&44.42&45.64&50.92\\
         \hline
         B + A  &86.87&63.9&70.33&37.13&27.56&67.29&33.91&54.95&44.93&73.23&28.52&17.28&26.95&73.34&62.09&64.11&56.44&55.81&41.44&42.23&51.42\\
         \hline
         B + S &86.9&64.45&70.74&36.8&27.39&67.04&32.63&54.34&45.35&72.91&28.02&15.6&26.28&74.01&62.25&64.45&59.3&57.84&45.55&46.62&51.93\\
         \hline
         B + A + S &87.22&63.64&71.35&38.95&31.33&68.08&34.13&55.07&46.47&73.15&28.74&20.28&24.11&74.12&63.12&65.38&57.39&56.96&42.49&42.52&52.22\\
         \hline
         B + S + A + L &87.04&63.29&70.48&40.06&31.69&68.29&39.89&55.37&49.06&73.17&31.49&22.44&25.59&73.4&62.23&65.14&59.1&58&44.21&44.14&53.05\\ 
         \hline
         \end{tabular}
         \label{tab:Ablation}
         }
         \end{table*}
   
Then we experiment our Smooth module on the baseline model, as shown in Tab.\ref{table:compare}, with the Smooth module added, mIou has increased by 1.01 points. This module also improves the accuracy of small object segmentation, the scores for the left and right parts are improved, this demonstrate that the Smooth module not only makes the model more robust to multi-scale objects, but also better combines low-level and high-level features. 

In the third experiment, we used the ASPP module together with the Smooth module, which is 1.3 percentage points higher than the baseline. In the last experiment of the ablation study, we added additional supervised loss to each output layer of Decoder, which not only makes the model more robust to multi-scale objects, but also adds semantic information to the features of Decoder output, helping model mix high and low features. As in Tab.\ref{table:compare}, multi-scale loss information helped the model increase by 0.8 in mIou, especially in the category of socks, and left-right part such as left-leg, right-leg. This means that the model not only classifies small objects more accurately but also has stronger semantic features.

\subsection{Comparison with state of the art}
\begin{figure}[h]
   \begin{center}
   \includegraphics[width=0.68\linewidth]{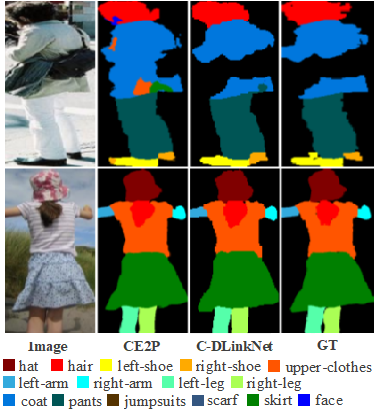}
   \end{center}
   \caption{Comparision of prediction on LIP validation set.}
   \label{fig:compare}
\end{figure}

We evaluate the performance of C-DLinkNet on the validation dataset of LIP and compare it to other state-of-the-art approaches. Without any bells and whistles, as shown in Tab.\ref{table:compare}, with almost half the input size, we get a comparable performance with CE2P\cite{CE2P_2018_CVPR} which input size is 473, and our method outperforms JPPNet by 1.68\% in terms of mIoU. It is worth noting that we did not use any additional annotation information. For example, as shown in the upper part of Fig.\ref{fig:compare}, we have better segmentation performance about coat than CE2P. 

\begin{table}[h]
   \caption{Results: C-DLinkNet can get better result. The input image size of JPP is 384*384 and CE2P uses size of 473.}
   \begin{center}
   \begin{tabular}{|l|c|c|c|}
   \hline
   Method  & pixel acc. & mean acc. & mIoU \\
   \hline
   Deeplab(VGG16)  & 82.66 & 51.64 & 41.64 \\
   Attention & 83.43 & 54.39 & 42.92 \\
   Deeplab(ResNet-101)\cite{Chen_2018_DE} & 84.09 & 55.62 & 44.8 \\
   JPPNet\cite{Liang2018Look} & 86.39 & 62.32 & 51.37 \\
   CE2P\cite{CE2P_2018_CVPR} & 87.37 & 63.20 & 53.10 \\
   C-DLinkNet & 87.04 & 62.84 & 53.05\\
   \hline
   \end{tabular}
   \end{center}
   \label{table:compare}   
   \end{table}

\section{Conclusion}
In this paper, we try to explore the differences and connections between human parsing and general semantic segmentation models, and use these features to build a model C-DLinkNet which effectively extracts with richer semantic information features and is more robust to multi-scale objects. We achieve comparable performance with smaller input size and no other external annotation information.

\bibliography{egbib}

\begin{thebibliography}{21}
\providecommand{\natexlab}[1]{#1}
\providecommand{\url}[1]{\texttt{#1}}
\expandafter\ifx\csname urlstyle\endcsname\relax
  \providecommand{\doi}[1]{doi: #1}\else
  \providecommand{\doi}{doi: \begingroup \urlstyle{rm}\Url}\fi

\bibitem[Berg et~al.(2012)Berg, Ortiz, Kiapour, and Yamaguchi]{Berg2012Parsing}
T.~L. Berg, L.~E. Ortiz, M.~H. Kiapour, and K.~Yamaguchi.
\newblock Parsing clothing in fashion photographs.
\newblock 2012.

\bibitem[Chen L~C(2018)]{Chen_2018_DE}
Papandreou~G Chen L~C, Zhu~Y.
\newblock Encoder-decoder with atrous separable convolution for semantic image
  segmentation.
\newblock In \emph{IEEE}, 2018.

\bibitem[Gong et~al.(2019)Gong, Gao, Liang, Shen, Wang, and
  Lin]{Gong_2019_CVPR}
Ke~Gong, Yiming Gao, Xiaodan Liang, Xiaohui Shen, Meng Wang, and Liang Lin.
\newblock Graphonomy: Universal human parsing via graph transfer learning.
\newblock In \emph{The IEEE Conference on Computer Vision and Pattern
  Recognition (CVPR)}, June 2019.

\bibitem[He et~al.(2015)He, Zhang, Ren, and Sun]{He2015Deep}
Kaiming He, Xiangyu Zhang, Shaoqing Ren, and Jian Sun.
\newblock Deep residual learning for image recognition.
\newblock 2015.

\bibitem[Ke et~al.(2017)Ke, Liang, Zhang, Shen, and Liang]{Ke2017Look}
Gong Ke, Xiaodan Liang, Dongyu Zhang, Xiaohui Shen, and Lin Liang.
\newblock Look into person: Self-supervised structure-sensitive learning and a
  new benchmark for human parsing.
\newblock 2017.

\bibitem[Krizhevsky et~al.(2012)Krizhevsky, Sutskever, and
  Hinton]{Krizhevsky2012ImageNet}
Alex Krizhevsky, I.~Sutskever, and G.~Hinton.
\newblock Imagenet classification with deep convolutional neural networks.
\newblock \emph{Advances in neural information processing systems}, 25\penalty0
  (2), 2012.

\bibitem[Li et~al.(2019)Li, M{\"{u}}ller, Thabet, and
  Ghanem]{DBLP:journals/corr/abs-1904-03751}
Guohao Li, Matthias M{\"{u}}ller, Ali~K. Thabet, and Bernard Ghanem.
\newblock Can gcns go as deep as cnns?
\newblock \emph{CoRR}, abs/1904.03751, 2019.
\newblock URL \url{http://arxiv.org/abs/1904.03751}.

\bibitem[Liang et~al.(2015)Liang, Xu, Shen, and Yang]{Liang2015Human}
Xiaodan Liang, Chunyan Xu, Xiaohui Shen, and Jianchao Yang.
\newblock Human parsing with contextualized convolutional neural network.
\newblock In \emph{IEEE International Conference on Computer Vision}, 2015.

\bibitem[Liang et~al.(2018)Liang, Ke, Shen, and Liang]{Liang2018Look}
Xiaodan Liang, Gong Ke, Xiaohui Shen, and Lin Liang.
\newblock Look into person: Joint body parsing and pose estimation network and
  a new benchmark.
\newblock \emph{IEEE Transactions on Pattern Analysis and Machine
  Intelligence}, 2018.

\bibitem[Long~J(2014)]{Long_2014}
Darrell~T Long~J, Shelhamer~E.
\newblock Fully convolutional networks for semantic segmentation.
\newblock In \emph{IEEE Transactions on Pattern Analysis and Machine
  Intelligence}, 2014.

\bibitem[Ruan(2018)]{CE2P_2018_CVPR}
Ruan.
\newblock Devil in the details- towards accurate single and multiple human
  parsing.
\newblock In \emph{the Association for the Advance of Artificial Intelligence},
  2018.

\bibitem[Sak et~al.()Sak, Senior, and Beaufays]{SakLong}
Haşim Sak, Andrew Senior, and Françoise Beaufays.
\newblock Long short-term memory based recurrent neural network architectures
  for large vocabulary speech recognition.

\bibitem[Simo-Serra et~al.(2014)Simo-Serra, Fidler, Moreno-Noguer, and
  Urtasun]{Simo2014A}
Edgar Simo-Serra, Sanja Fidler, Francesc Moreno-Noguer, and Raquel Urtasun.
\newblock \emph{A High Performance CRF Model for Clothes Parsing}.
\newblock 2014.

\bibitem[University()]{lip_web}
Sun Yat-Sen University.
\newblock Lip, look into person.
\newblock \url{http://sysu-hcp.net/lip/index.php}.
\newblock Accessed May 24, 2019.

\bibitem[Wang et~al.(2019)Wang, Zhang, Qi, Shen, Pang, and
  Shao]{Wang_2019_ICCV}
Wenguan Wang, Zhijie Zhang, Siyuan Qi, Jianbing Shen, Yanwei Pang, and Ling
  Shao.
\newblock Learning compositional neural information fusion for human parsing.
\newblock In \emph{The IEEE International Conference on Computer Vision
  (ICCV)}, October 2019.

\bibitem[Xie et~al.(2016)Xie, Girshick, Dollár, Tu, and He]{Xie2016}
Saining Xie, Ross Girshick, Piotr Dollár, Zhuowen Tu, and Kaiming He.
\newblock Aggregated residual transformations for deep neural networks.
\newblock \emph{arXiv preprint arXiv:1611.05431}, 2016.

\bibitem[Xu(2020)]{xu2020deep}
Peng Xu.
\newblock Deep learning for free-hand sketch: A survey.
\newblock \emph{arXiv preprint arXiv:2001.02600}, 2020.

\bibitem[Xu et~al.(2019)Xu, Joshi, and Bresson]{xu2019multigraph}
Peng Xu, Chaitanya~K Joshi, and Xavier Bresson.
\newblock Multi-graph transformer for free-hand sketch recognition.
\newblock \emph{arXiv preprint arXiv:1912.11258}, 2019.

\bibitem[Zhao et~al.(2017{\natexlab{a}})Zhao, Shi, Qi, Wang, and
  Jia]{Zhao_2017_CVPR}
Hengshuang Zhao, Jianping Shi, Xiaojuan Qi, Xiaogang Wang, and Jiaya Jia.
\newblock Pyramid scene parsing network.
\newblock In \emph{The IEEE Conference on Computer Vision and Pattern
  Recognition (CVPR)}, 2017{\natexlab{a}}.

\bibitem[Zhao et~al.(2017{\natexlab{b}})Zhao, Li, Nie, Zhao, Chen, Wang, Feng,
  and Yan]{Zhao_2017_CVPR_Workshops}
Jian Zhao, Jianshu Li, Xuecheng Nie, Fang Zhao, Yunpeng Chen, Zhecan Wang,
  Jiashi Feng, and Shuicheng Yan.
\newblock Self-supervised neural aggregation networks for human parsing.
\newblock In \emph{The IEEE Conference on Computer Vision and Pattern
  Recognition (CVPR) Workshops}, July 2017{\natexlab{b}}.

\bibitem[Zhou et~al.(2018)Zhou, Zhang, and Wu]{Zhou_2018_CVPR_Workshops}
Lichen Zhou, Chuang Zhang, and Ming Wu.
\newblock D-linknet: Linknet with pretrained encoder and dilated convolution
  for high resolution satellite imagery road extraction.
\newblock In \emph{The IEEE Conference on Computer Vision and Pattern
  Recognition (CVPR) Workshops}, June 2018.

\end{thebibliography}
\end{document}